\documentclass[conference]{IEEEtran}
\usepackage{amsmath,amsfonts}
\usepackage{algorithmic}
\usepackage{algorithm}
\usepackage{array}
\usepackage[caption=false,font=normalsize,labelfont=sf,textfont=sf]{subfig}
\usepackage{textcomp}
\usepackage{stfloats}
\usepackage{url}
\usepackage{verbatim}
\usepackage{graphicx}
\usepackage{cite}
\begin{document}

\title{PyCAT4: A Hierarchical Vision Transformer-based Framework \\ for 3D Human Pose Estimation}
\author{
\IEEEauthorblockN{Zongyou~Yang}
\IEEEauthorblockA{
Department of Computer Science\\
University College London\\
London, United Kingdom\\
dryang0624@gmail.com
}
\and
\IEEEauthorblockN{Jonathan~Loo}
\IEEEauthorblockA{
School of Electronic Engineering\\
and Computer Science\\
Queen Mary University of London\\
London, United Kingdom\\
j.loo@qmul.ac.uk
}
\and
\IEEEauthorblockN{Yinghan~Hou}
\IEEEauthorblockA{
Department of Earth Science\\
and Engineering\\
Imperial College London\\
London, United Kingdom\\
ghwzhyinghan@gmail.com
}
}




\maketitle

\begin{abstract}
Recently, a significant improvement in the accuracy of 3D human pose estimation has been achieved by combining convolutional neural networks (CNNs) with pyramid grid alignment feedback loops. Additionally, innovative breakthroughs have been made in the field of computer vision through the adoption of Transformer-based temporal analysis architectures. Given these advancements, this study aims to deeply optimize and improve the existing Pymaf network architecture. The main innovations of this paper include: (1) Introducing a Transformer feature extraction network layer based on self-attention mechanisms to enhance the capture of low-level features; (2) Enhancing the understanding and capture of temporal signals in video sequences through feature temporal fusion techniques; (3) Implementing spatial pyramid structures to achieve multi-scale feature fusion, effectively balancing feature representations differences across different scales. The new PyCAT4 model obtained in this study is validated through experiments on the COCO and 3DPW datasets. The results demonstrate that the proposed improvement strategies significantly enhance the network's detection capability in human pose estimation, further advancing the development of human pose estimation technology.
\end{abstract}

\begin{IEEEkeywords}
Human pose estimation, Deep Learning, Transformer, Attention Mechanism
\end{IEEEkeywords}

\section{Introduction}
\begin{figure}[!t]
\centering
\includegraphics[width=2.5in]{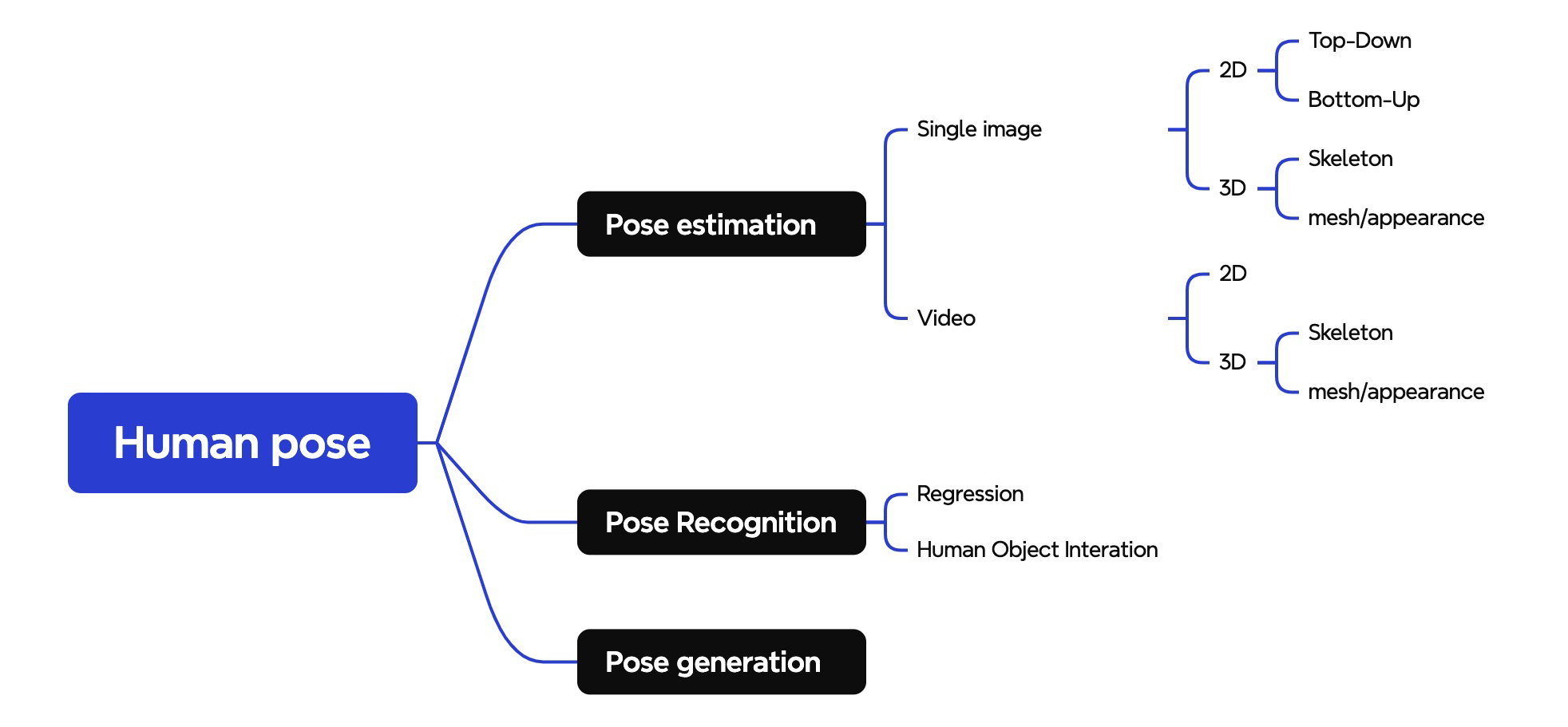}
\caption{Classification of Human Pose Estimation}
\label{fig:hpe-classification}
\end{figure}

\IEEEPARstart{H}{uman posture recognition} encompasses a multifaceted domain within computer vision, which is currently divided into three principal research categories as depicted in Figure~\ref{fig:hpe-classification}:

\begin{enumerate}
    \item Human Pose Estimation (HPE) constitutes a fundamental challenge in computer vision and serves as the groundwork for numerous higher-level semantic tasks and downstream applications. It essentially entails the estimation of the human body's posture by identifying key points such as the head, left hand, right foot, etc.
    \item Human Action Recognition, as opposed to the mere measurement tasks of pose estimation, focuses more on semantic understanding and is considered a downstream semantic classification task. It is typically built upon the foundation provided by pose estimation.
    \item Human Pose Generation involves the creation of virtual actions through cross-modal or action synthesis approaches, representing a relatively emerging field~\cite{zheng2023deep}.
\end{enumerate}

HPE tasks can be categorized into single-frame and temporal estimation. Single-frame estimation predicts keypoints from a 2D RGB image, while temporal estimation incorporates time-flow to model pose in video sequences. Output can also be divided into 2D skeleton estimation and 3D mesh recovery.

\begin{figure}[!t]
\centering
\includegraphics[width=2.5in]{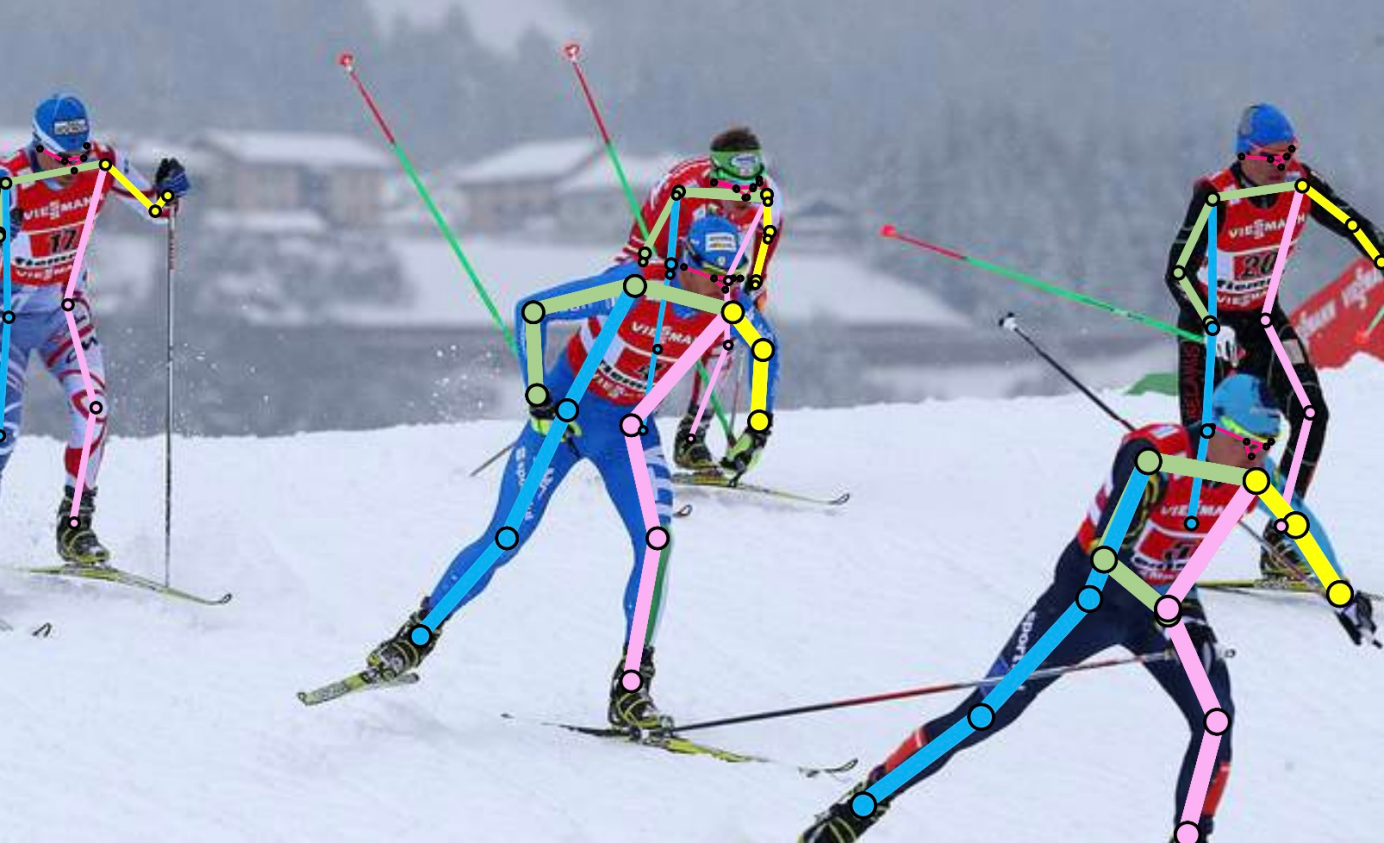} 
\caption{2D Human Pose Estimation~\cite{wang2020deep}}
\label{fig:pose2d}
\end{figure}

\begin{figure}[!t]
\centering
\includegraphics[width=2.5in]{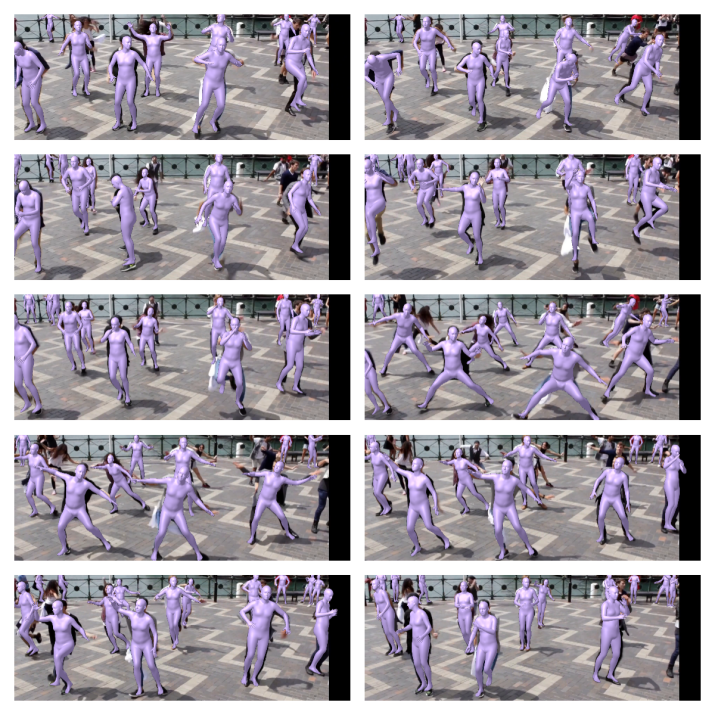} 
\caption{3D Human Pose Estimation}
\label{fig:pose3d}
\end{figure}

Applications of HPE span gaming, VR, sports, security, and healthcare. For instance, in sports, it aids in athlete performance analysis, while in healthcare it supports rehabilitation assessment.

\subsection{Problem Statement}
Despite progress in 3D HPE using CNNs, pyramid alignment feedback, and Transformer architectures, key challenges remain:
\begin{enumerate}
    \item The estimation of posture in single images does not accurately capture the dynamic continuity of human body movement.
    \item Current models exhibit insufficient utilization of temporal information when processing video sequence data, leading to diminished coherence and accuracy in pose estimation.
    \item Multi-scale feature fusion technology has not been fully utilized, limiting the ability of the model to extract features at different levels.
    \item The real-time performance and efficiency of existing technologies still need to be improved, limiting their wide application in the real world.
\end{enumerate}

\subsection{Purpose of the Study}
This study aims to solve the problem of human 3D pose estimation by introducing the latest deep learning architecture. Firstly, Transform feature extraction layer network structure is combined with self-attention mechanism, feature space-time fusion technology and multi-scale feature fusion strategy. To improve the accuracy, consistency and real-time performance of human pose estimation model from single image and video sequence, this work aims to further promote the development of human pose estimation technology and provide technical support for the application in related fields.

\subsection{Objectives}
\begin{itemize}
    \item Design Transformer-based feature extraction for enhanced spatial encoding.
    \item Develop spatio-temporal fusion to utilize motion data.
    \item Apply spatial pyramid fusion for multi-scale feature alignment.
    \item Evaluate on COCO and benchmark against prior methods.
    \item Implement a real-time 3D HPE system.
\end{itemize}

\subsection{Research Questions}
\begin{enumerate}
    \item How does self-attention improve Transformer-based 3D HPE?
    \item Can spatio-temporal fusion increase accuracy in video-based estimation?
    \item What is the contribution of multi-scale fusion to estimation performance?
    \item How does the proposed method compare in accuracy and efficiency?
    \item How can real-time estimation be achieved effectively?
\end{enumerate}

\section{Related Work}
Our work builds upon the PyMAF framework~\cite{zhang2021pymaf}, extending it with several recent advances in vision modeling. Specifically, we integrate Coordinate Attention~\cite{hou2021coordinate}, a hierarchical Swin Transformer backbone~\cite{liu2021swin}, and spatial-temporal Transformer-based fusion inspired by PoseFormer~\cite{zheng20213d} to construct a unified architecture for 3D human pose estimation.

In contrast to PoseFormer~\cite{zheng20213d}, which primarily addresses temporal modeling in videos, our proposed PyCAT4 introduces a multi-scale fusion module (FPN~\cite{lin2017feature} + ASPP~\cite{he2014spatial}) and spatial attention mechanisms to jointly enhance both temporal coherence and spatial detail representation.

While PyMAF~\cite{zhang2021pymaf} and PoseFormer~\cite{zheng20213d} provide strong baselines in their respective domains, to the best of our knowledge, the combined use of these modules within a single pipeline for real-time and accurate 3D pose estimation is first proposed in this work.

\subsection{Theoretical Framework}

Human pose estimation pipelines based on deep learning are generally categorized into two major approaches: regression-based methods and heatmap-based methods. Regression methods utilize end-to-end architectures to directly map input images to joint locations or body model parameters. In contrast, heatmap-based methods focus on predicting body part locations via heatmap representations.

\subsubsection{Regression Methods}
\begin{figure}[!t]
\centering
\includegraphics[width=2.5in]{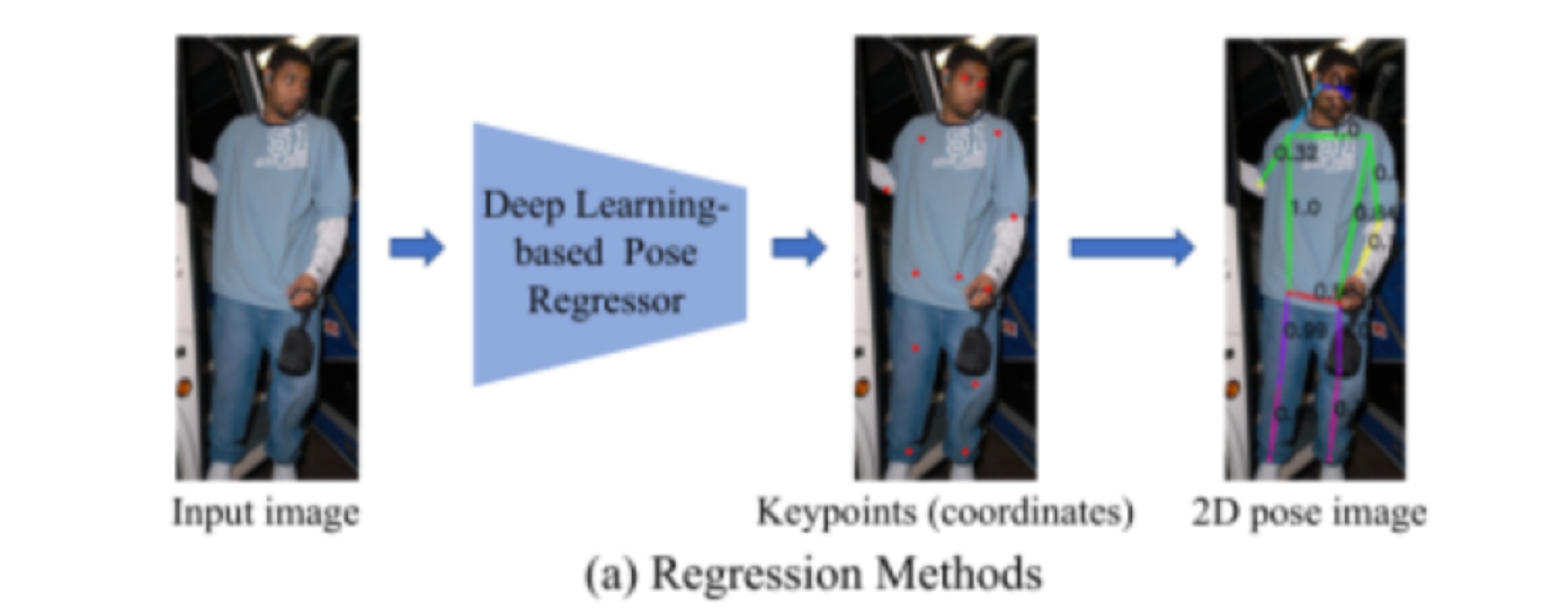}
\caption{Illustration of the regression-based human pose estimation method~\cite{zheng2023deep}.}
\label{fig_regression}
\end{figure}

Regression methods directly predict joint coordinates from raw image inputs. As shown in Fig.~\ref{fig_regression}, the approach typically involves feature extraction via a convolutional neural network (CNN), followed by a regression head that outputs joint positions in the image space.

\subsubsection{Heatmap-Based Methods}
\begin{figure}[!t]
\centering
\includegraphics[width=2.5in]{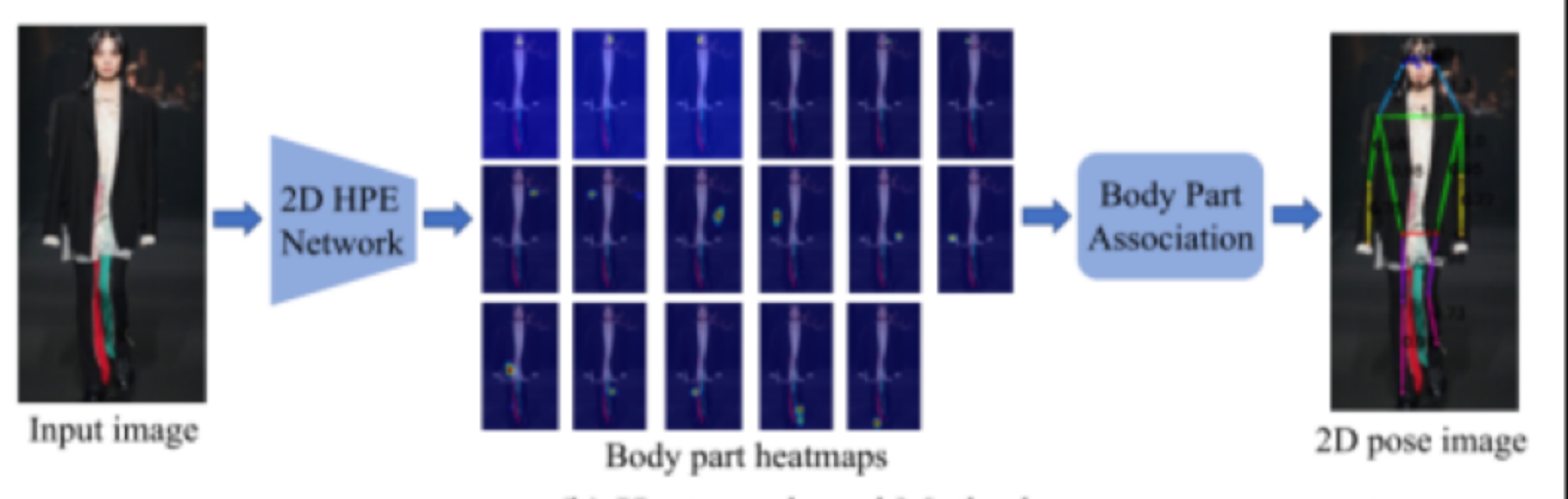}
\caption{Illustration of the heatmap-based method~\cite{zheng2023deep}.}
\label{fig_heatmap}
\end{figure}

Instead of directly regressing keypoints, heatmap-based methods generate 2D heatmaps centered around each joint. These methods leverage spatial context and Gaussian-based supervision to stabilize training and improve accuracy. Fig.~\ref{fig_heatmap} illustrates this approach.

\subsubsection{3D Human Pose Estimation}
\begin{figure}[!t]
\centering
\includegraphics[width=3in]{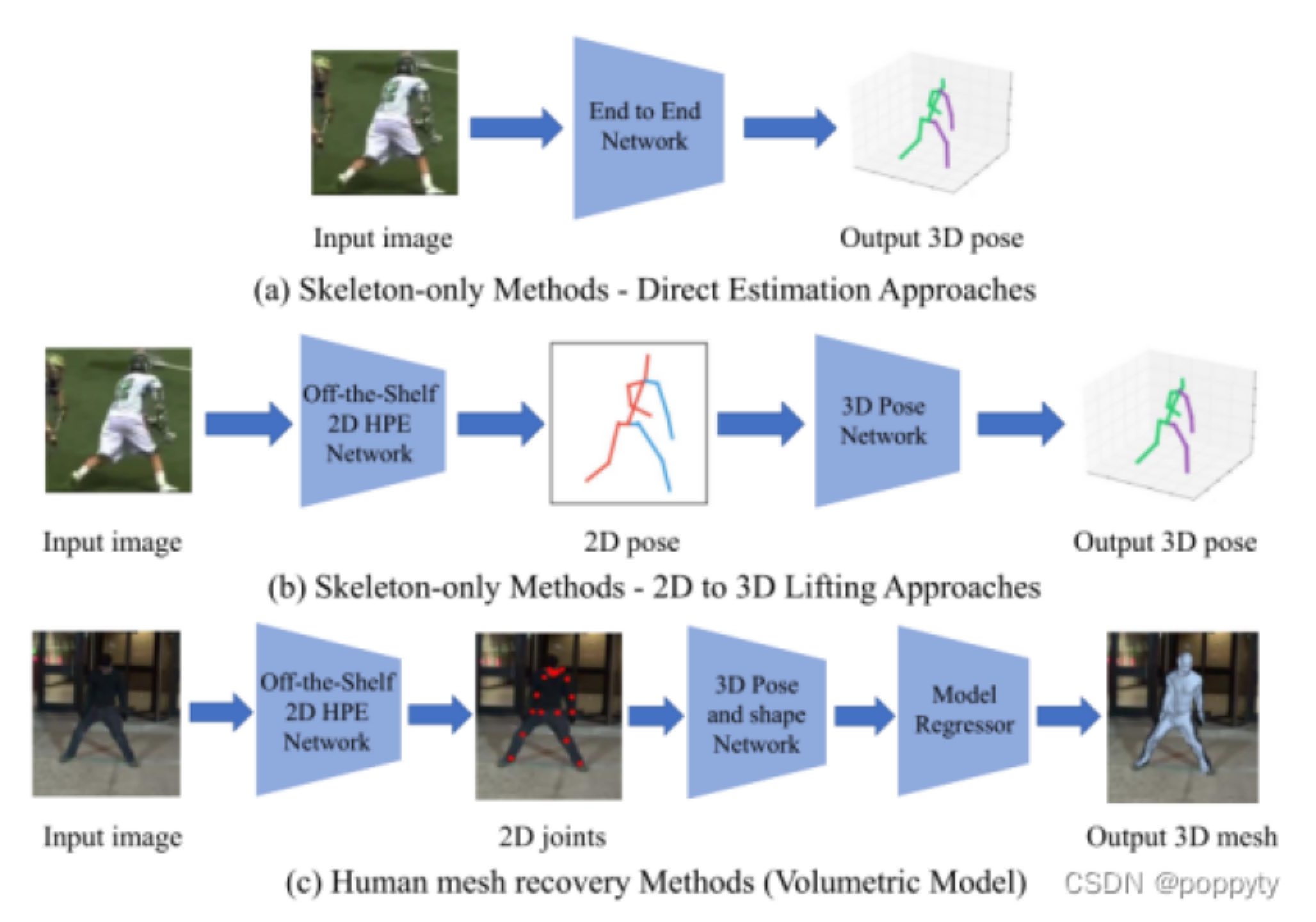}
\caption{Frameworks for 3D human pose estimation~\cite{zheng2023deep}. (a) Direct estimation from image, (b) 2D-to-3D lifting, (c) Mesh recovery via model regression.}
\label{fig_3dpose}
\end{figure}

Recent interest in 3D human pose estimation stems from its ability to provide spatially rich representations of the human body. Unlike 2D estimation, the 3D variant outputs either joint coordinates in three-dimensional space or detailed mesh structures. As illustrated in Fig.~\ref{fig_3dpose}, these methods can be grouped into three categories: direct 3D estimation from images, 2D-to-3D lifting strategies, and full mesh reconstruction using parametric models.

\subsection{Concept Definitions}

\subsubsection{Pose Estimation}
\textbf{2D Pose Estimation:} Infers pixel-level joint positions from single-view images, typically used in simpler applications with lower spatial requirements.

\textbf{3D Pose Estimation:} Reconstructs human posture in three-dimensional space from images or video. It provides more comprehensive information, making it suitable for applications such as motion analysis and virtual reality.

\subsubsection{Keypoints and Skeleton}
\textbf{Keypoints:} Critical anatomical landmarks on the human body, such as head, shoulders, elbows, and knees, used to represent posture.

\textbf{Skeleton:} A graphical structure that connects keypoints to illustrate the body’s shape and posture.

\subsubsection{Deep Learning Models}
\textbf{CNNs:} Convolutional neural networks effectively learn spatial hierarchies from visual data and have been foundational in human pose estimation~\cite{he2015deep}.

\textbf{Transformers:} Originally developed for NLP, Transformers have demonstrated strong performance in vision tasks, leveraging self-attention to model long-range dependencies~\cite{vaswani2017attention}.

\subsubsection{Self-Attention Mechanism}
This mechanism enables a model to assign differential importance to various elements in a sequence, aiding in refined focus on key regions relevant to pose estimation~\cite{guo2022attention}.

\subsubsection{Spatial-Temporal Transformer}
This module models spatio-temporal dependencies by encoding both frame-wise features and their temporal dynamics. It enhances both spatial localization and motion continuity~\cite{zheng20213d}.

\subsection{Advances in Deep Learning for Pose Estimation}

\subsubsection{Latest Models and Techniques}
State-of-the-art deep learning models, including CNNs and Transformers, excel in extracting hierarchical and contextual features from images and videos, significantly improving pose estimation performance.

\subsubsection{Self-Attention and Transformer Models}
Self-attention mechanisms enable focus on critical spatial and temporal regions. When applied to pose estimation, these mechanisms improve robustness in dynamic and multi-person scenes.

\subsubsection{Multi-Scale Feature Fusion and Temporal Analysis}
Integrating multi-scale feature fusion allows models to capture both local details and global context. Temporal analysis techniques enhance consistency and enable motion prediction~\cite{lin2017feature}.

\subsubsection{Practical Application Case Studies}
Human pose estimation has found application in sports analytics, virtual reality, health monitoring, and security. These domains benefit from real-time posture analysis and interactive capabilities enabled by deep learning.

\section{Proposed Method}

\subsection{Research Design}
This study uses the PyMAF model as a benchmark to systematically evaluate the effectiveness of various deep network modules, including coordinate attention mechanisms, Transformer-based architectures, temporal fusion strategies, and multi-scale feature aggregation. To ensure rigorous experimental comparisons and uphold scientific validity, an ablation study methodology was employed under strict control of variables. The sequential design and implementation steps are as follows:

\subsubsection{Baseline Model}
The baseline configuration is established by training the original PyMAF model without any architectural modifications. This model serves as a control reference for all subsequent enhancements~\cite{zhang2021pymaf}.

\subsubsection{Integration of Coordinate Attention (CA)}
Coordinate Attention modules are incorporated into the baseline’s feature extraction layer. All other settings remain unchanged to isolate the effect of CA on model performance~\cite{hou2021coordinate}.

\subsubsection{Introduction of Swin Transformer}
The model is further enhanced by replacing its CNN backbone with a Swin Transformer, recognized for its capacity to model hierarchical features in images~\cite{liu2021swin}. Training settings are held constant to evaluate the direct impact of this substitution.

\subsubsection{Multi-Scale Fusion Implementation}
A multi-scale feature fusion module is introduced to aggregate information across different spatial resolutions and semantic depths, refining model performance~\cite{lin2017feature}.

\subsubsection{Temporal Fusion Integration}
Inspired by PoseFormer~\cite{zheng20213d}, a temporal fusion module is incorporated to exploit temporal dependencies in video sequences. Its contribution is assessed against prior models under identical training conditions.

\subsubsection{Controlled Experimental Conditions}
To ensure fair comparisons, only one variable is altered in each experimental setup. This strategy guarantees the objectivity and reproducibility of performance evaluations.

\subsection{Data Collection and Preparation}
Two datasets are used: COCO2014 and 3DPW.

\begin{itemize}
    \item \textbf{COCO2014:} A widely used benchmark in computer vision containing over 330,000 annotated images across diverse scenarios. It is employed for both training and validation.
    \item \textbf{3DPW:} A dataset comprising outdoor video sequences annotated with accurate 3D poses, collected via mobile devices, supporting evaluation in natural environments.
\end{itemize}

These datasets are used as a mixed training set, and their validation splits are employed for evaluation.

\subsection{Model Architecture and Implementation}
The PyMAF architecture consists of three major modules, illustrated in Fig.~\ref{fig_pyamf}:

\begin{figure}[!t]
\centering
\includegraphics[width=3in]{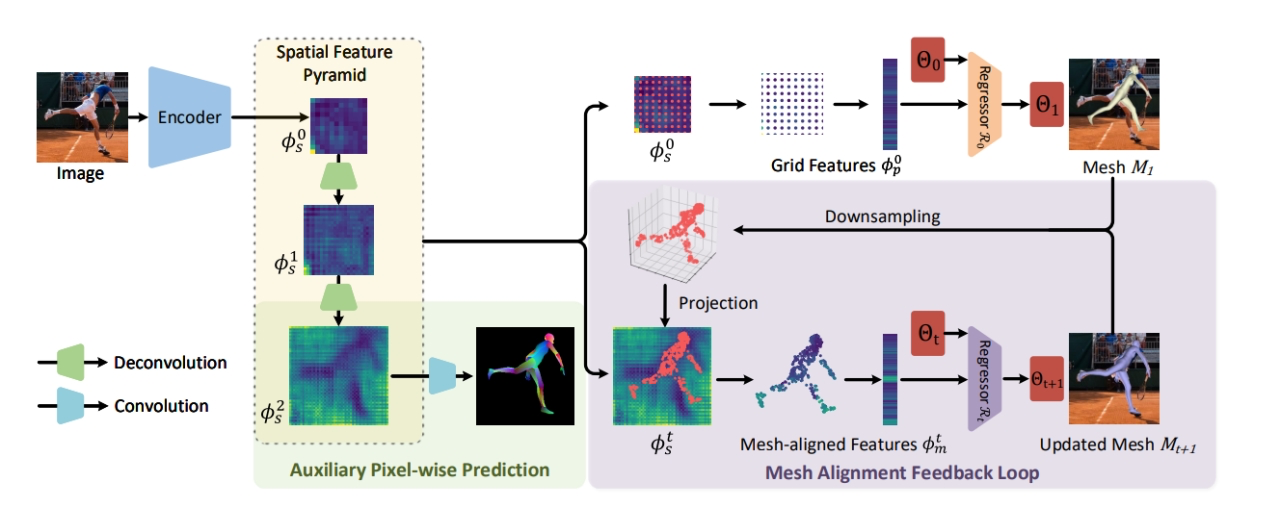}
\caption{Overview of the PyMAF architecture~\cite{zhang2021pymaf}.}
\label{fig_pyamf}
\end{figure}

\begin{itemize}
    \item \textbf{Feature Extraction Layer:} Utilizes ResNet-50 with four downsampling stages to capture both low-level and high-level features. Final feature map resolution is reduced by a factor of 32.
    \item \textbf{Auxiliary Per-Pixel Prediction:} Uses deconvolution layers to upsample features for dense supervision, improving robustness against occlusion and lighting variations.
    \item \textbf{Mesh Alignment Feedback Loop:} Based on iterative error feedback (IEF), this loop refines pose parameters using global features and current estimates to iteratively reduce mesh alignment errors \cite{zhang2021pymaf}.
\end{itemize}

\subsection{Training Strategy}

\begin{itemize}
    \item Backbone: ResNet-50 pretrained on ImageNet; input size 224x224; output features 2048\(\times\)7\(\times\)7.
    \item Global feature extraction: Average pooling yields 2048\(\times\)1 vector for the HMR module.
    \item Deconvolution: Spatial feature maps of sizes \{14\(\times\)14, 28\(\times\)28, 56\(\times\)56\} are generated.
    \item Optimizer: Adam with a learning rate of 5e-5, batch size 128.
    \item Hardware: Trained on six NVIDIA RTX 4090 GPUs for 60 epochs.
    \item Learning Rate: No decay applied.
\end{itemize}

\subsection{Evaluation Metrics}

\subsubsection{3D Evaluation Metrics (3DPW/COCO)}

\textbf{PVE (Per-Vertex Error):} Measures average Euclidean distance between predicted and ground truth mesh vertices, reported in millimeters.

\textbf{MPJPE (Mean Per Joint Position Error):} Calculates the mean Euclidean distance between predicted and actual 3D joint positions.

\textbf{PA-MPJPE (Procrustes Aligned MPJPE):} MPJPE after rigid alignment to account for scale, rotation, and translation differences.

\subsubsection{2D Evaluation Metrics (COCO)}

\textbf{Average Precision (AP):} Evaluates model precision across recall levels using Object Keypoint Similarity (OKS).

\textbf{Average Recall (AR):} Measures recall over varying OKS thresholds.

\textbf{OKS:} A similarity metric analogous to IoU, accounting for keypoint localization error, scale, and visibility.

\textbf{AP$_{50}$:} AP at OKS threshold 0.5.

\textbf{AP$_{50:95}$:} Mean AP across OKS thresholds from 0.5 to 0.95 in 0.05 increments.

These metrics provide a comprehensive and robust evaluation of pose estimation models across both 2D and 3D benchmarks.


\section{Design and Implementation}
This chapter details the design and implementation of our proposed model, PyCAT4. We begin by defining the baseline architecture and then progressively introduce a series of enhancements, including attention mechanisms, a Transformer-based backbone, multi-scale feature fusion, and temporal analysis modules.

\subsection{Baseline Model}
\begin{figure}[!t]
    \centering
    \includegraphics[width=3in]{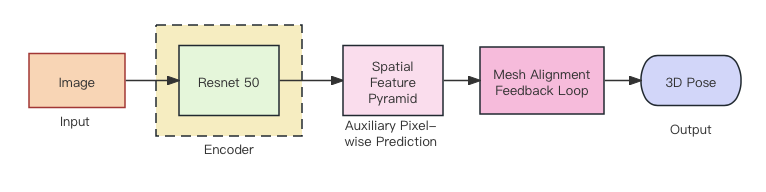} 
    \caption{The architecture of the baseline PyMAF model, which includes a feature pyramid, an auxiliary pixel-wise prediction head, and a mesh alignment feedback loop \cite{zhang2021pymaf}.}
    \label{fig:pymaf_arch}
\end{figure}

Our baseline is the original PyMAF network structure ~\cite{zhang2021pymaf}, which serves as a benchmark for evaluating all subsequent architectural improvements. As illustrated in Fig.~\ref{fig:pymaf_arch}, PyMAF employs a deep regressor with a pyramidal mesh alignment feedback loop to estimate 3D human pose and shape from a single image. This model provides a strong foundation but has limitations in capturing global context and fine-grained details, which we aim to address.

\subsection{Attention Mechanisms}
\begin{figure}[!t]
    \centering
    \includegraphics[width=2.5in]{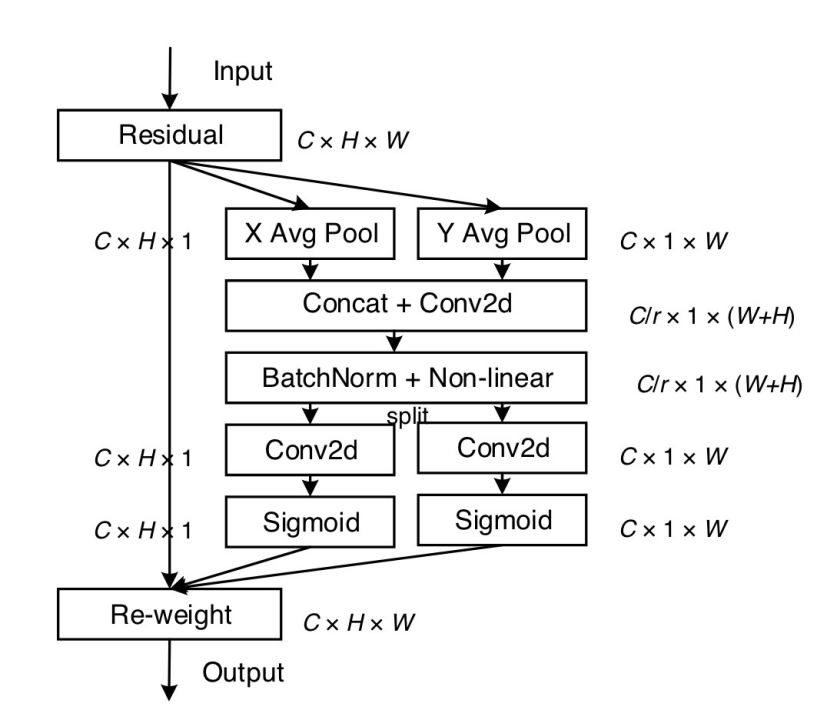}
    \caption{Schematic diagram of the Coordinate Attention (CA) module architecture. It decomposes channel attention into two 1D feature encoding processes to capture spatial information along two directions \cite{hou2021coordinate}.}
    \label{fig:ca_arch}
\end{figure}

Attention mechanisms in computer vision enhance model performance by enabling them to selectively focus on the most informative regions of an input image \cite{hu2021attention_survey}. To improve the feature extraction capability of our network, we introduce the Coordinate Attention (CA) module, a lightweight and efficient mechanism designed to capture not only inter-channel relationships but also direction-aware and position-sensitive information \cite{hou2021coordinate}. A schematic of the CA module is provided in Fig.~\ref{fig:ca_arch}.

\subsubsection{Implementation and Integration of the CA Module}
\begin{figure}[!t]
    \centering
    \includegraphics[width=\columnwidth]{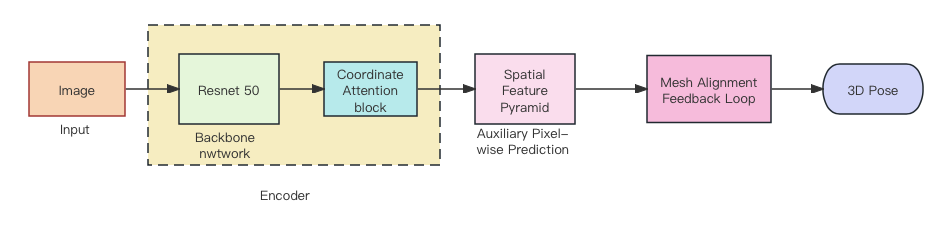}
    \caption{The model architecture after integrating the Coordinate Attention (CA) block into the backbone network.}
    \label{fig:ca_integration}
\end{figure}

The CA module is strategically embedded within the feature extraction backbone, immediately following the deep feature extraction stages of the ResNet-50. This placement allows the module to refine crucial features before they are passed to subsequent layers. As shown in Fig.~\ref{fig:ca_integration}, the CA block seamlessly integrates into the architecture without altering the dimensionality of the feature maps, preserving the data flow.

\subsubsection{Contribution and Validation}
By coordinating attention across spatial dimensions, the CA module efficiently captures both local and global contextual information, which is particularly beneficial for estimating pose in complex scenes. The efficacy of this module is rigorously validated through comparative ablation studies on the COCO and 3DPW datasets, where its performance is quantified against the baseline model.

\subsection{Swin Transformer Backbone}
\begin{figure}[!t]
    \centering
    \includegraphics[width=\columnwidth]{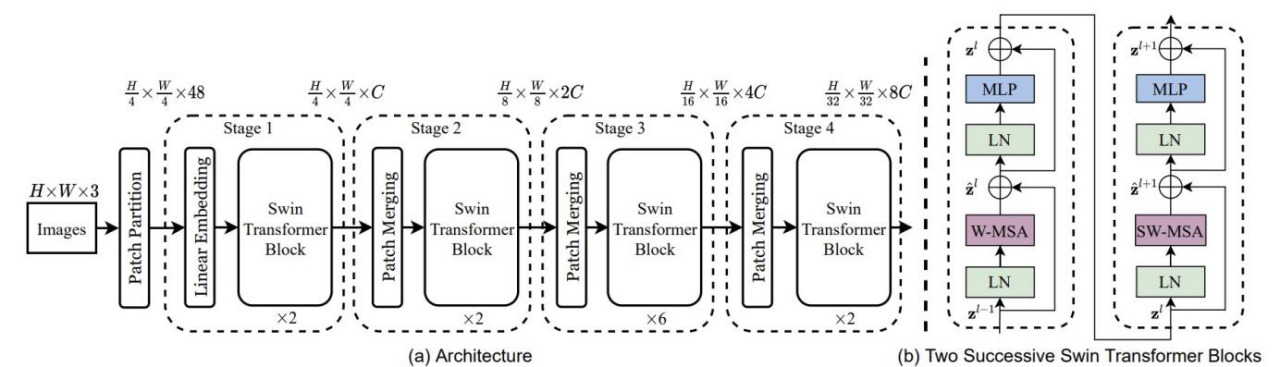}
    \caption{The architecture of the Swin Transformer, showing its hierarchical design with patch merging and successive Swin Transformer blocks \cite{liu2021swin}.}
    \label{fig:swin_arch}
\end{figure}

\begin{figure}[!t]
    \centering
    \includegraphics[width=\columnwidth]{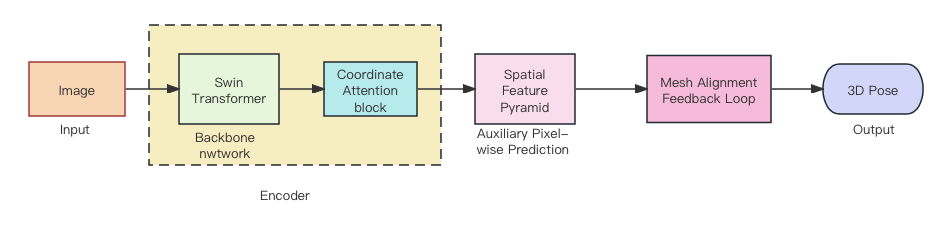}
    \caption{The model architecture with the ResNet backbone replaced by the Swin Transformer.}
    \label{fig:swin_integration}
\end{figure}

To enhance the model's ability to capture complex patterns and long-range dependencies, we replace the conventional ResNet backbone with a Swin Transformer \cite{liu2021swin}. The Swin Transformer introduces a hierarchical structure and a shifted window multi-head self-attention (W-MSA/SW-MSA) mechanism, which provides greater efficiency and modeling power for vision tasks. The architecture is depicted in Fig.~\ref{fig:swin_arch}. This substitution, illustrated in Fig.~\ref{fig:swin_integration}, is a core component of our PyCAT4 model. The effectiveness of this change is validated through ablation studies on the COCO and 3DPW datasets. 

\subsection{Multi-scale Feature Fusion}
\begin{figure}[!t]
    \centering
    \subfloat[]{\includegraphics[width=1.6in]{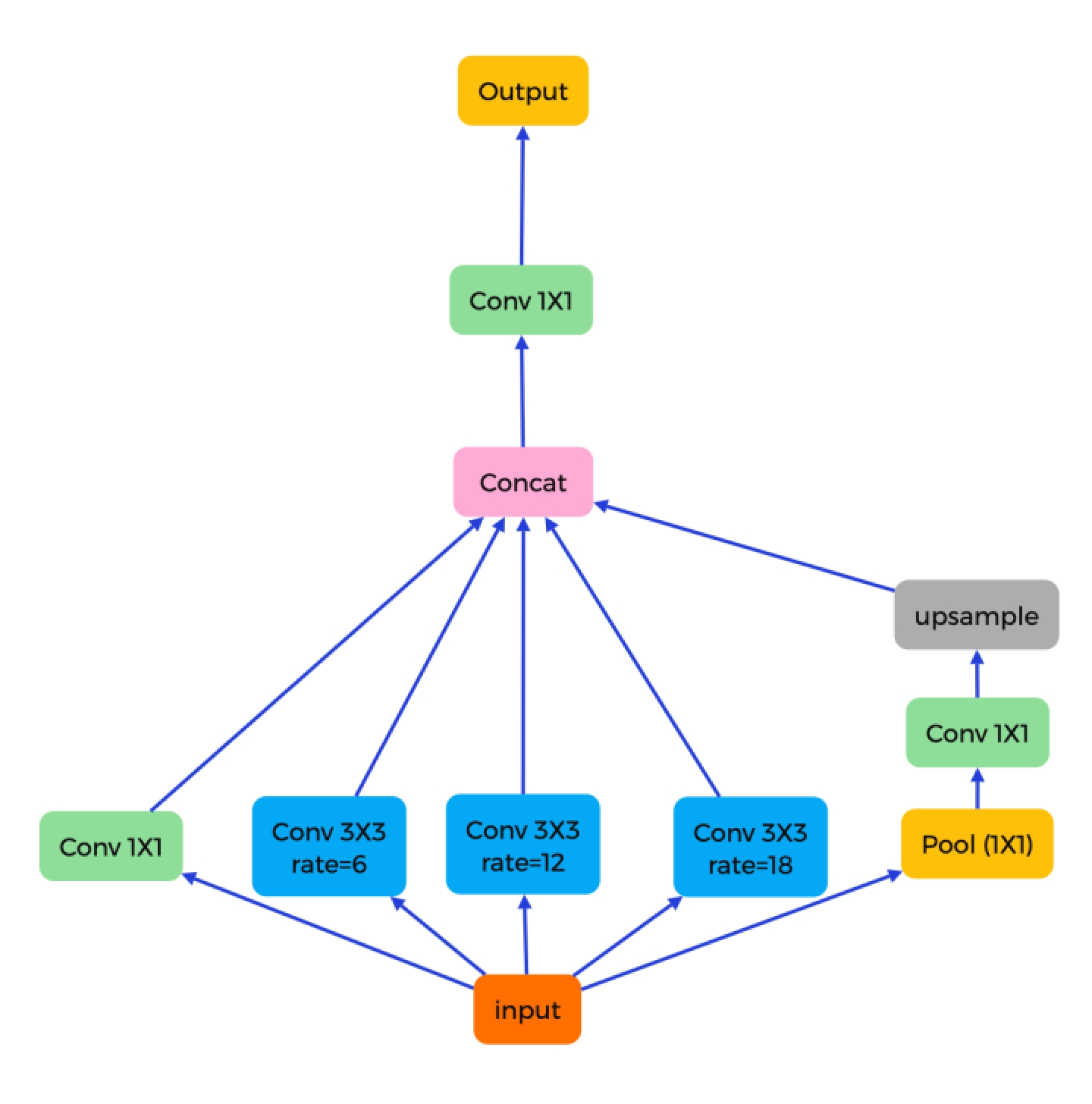}\label{fig:aspp_arch}}
    \hfil
    \subfloat[]{\includegraphics[width=1.6in]{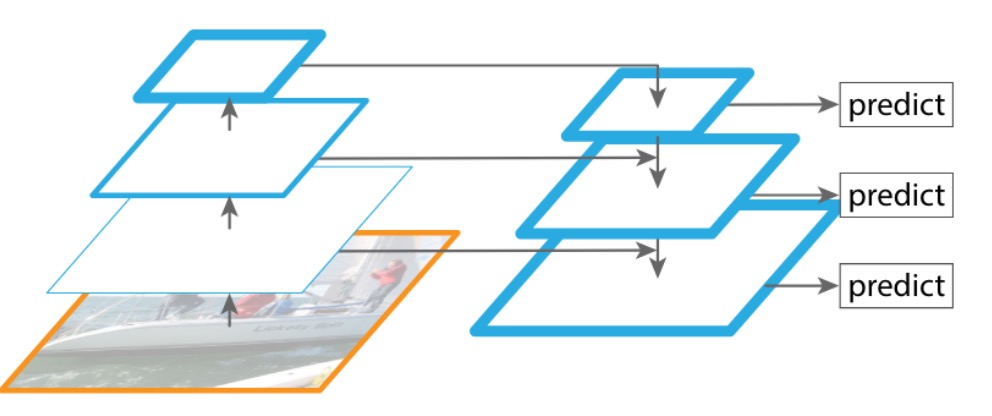}\label{fig:fpn_arch}}
    \caption{(a) Architecture of the Atrous Spatial Pyramid Pooling (ASPP) module. (b) Architecture of the Feature Pyramid Network (FPN).}
\end{figure}

\begin{figure}[!t]
    \centering
    \includegraphics[width=\columnwidth]{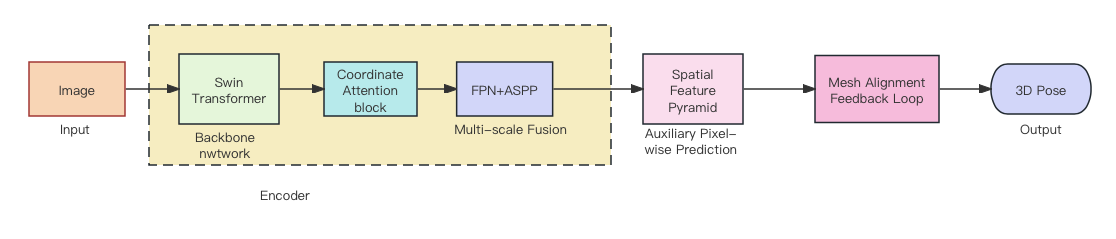}
    \caption{The backbone architecture incorporating the FPN+ASPP multi-scale fusion module.}
    \label{fig:fpn_aspp_integration}
\end{figure}

To enhance the model's ability to interpret features at various scales, we implement an advanced multi-scale fusion strategy \cite{lin2017feature}. Specifically, we integrate an Atrous Spatial Pyramid Pooling (ASPP) structure, renowned for its effectiveness in capturing multi-scale context without resolution loss, with a Feature Pyramid Network (FPN). ASPP uses dilated convolutions with different rates to expand the receptive field (Fig.~\ref{fig:aspp_arch}), while FPN combines low-resolution, semantically strong features with high-resolution, semantically weak features (Fig.~\ref{fig:fpn_arch}). This combined FPN+ASPP module, shown in Fig.~\ref{fig:fpn_aspp_integration}, ensures that both high-level semantic cues and low-level positional details are effectively captured and utilized.

\subsection{Temporal Fusion Module}
\begin{figure}[!t]
    \centering
    \includegraphics[width=\columnwidth]{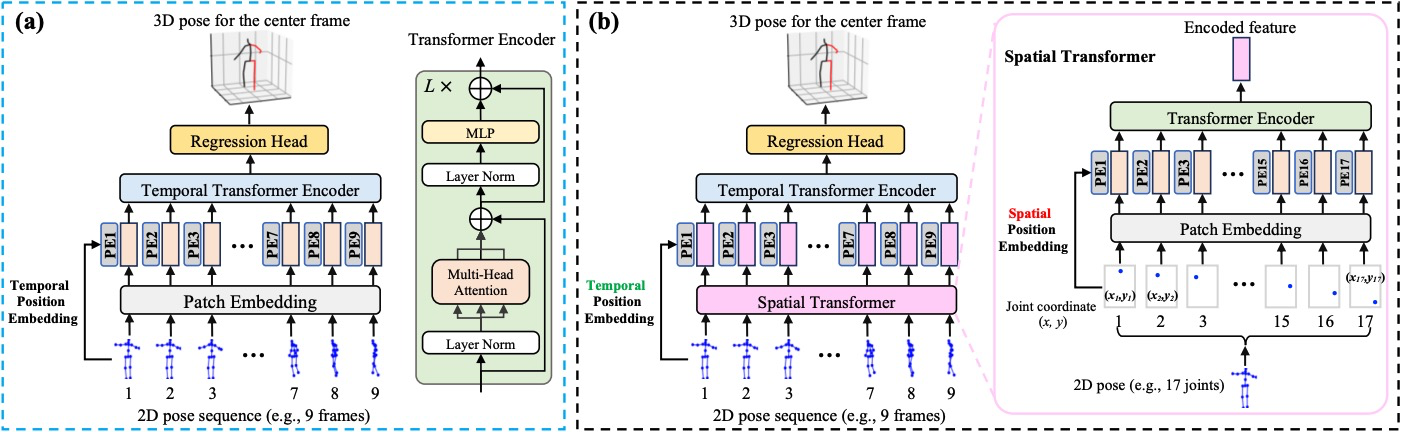}
    \caption{The Spatial-Temporal Transformer architecture proposed in PoseFormer, which serves as the inspiration for our temporal fusion module \cite{zheng20213d}.}
    \label{fig:poseformer_arch}
\end{figure}

\begin{figure}[!t]
    \centering
    \includegraphics[width=2in]{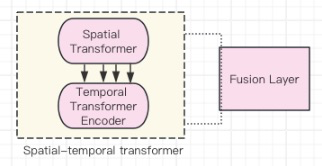}
    \caption{Architecture of our integrated Temporal Fusion network.}
    \label{fig:temporal_arch}
\end{figure}

For video-based pose estimation, we introduce a temporal fusion module to synthesize feature information from both current and preceding frames \cite{zheng20213d}. Traditional single-frame methods often fail to capture the fluidity of human motion. By integrating temporal data, our model processes a richer, time-extended context, achieving a more natural and accurate reconstruction of motion. We leverage a Transformer-based architecture, inspired by PoseFormer (Fig.~\ref{fig:poseformer_arch}), to model these extensive time dependencies. The final architecture of this module is shown in Fig.~\ref{fig:temporal_arch}.

\subsection{Real-time Detection System}
To demonstrate practical applicability, we developed a real-time detection system by integrating our trained models with the OpenCV library. The system captures live video feeds, performs frame-by-frame inference on a GPU, and displays the estimated 3D pose in real-time, making it suitable for applications in interactive entertainment and motion analysis.

\subsection{The Proposed PyCAT4 Architecture}
\begin{figure*}[!t]
    \centering
    \includegraphics[width=0.9\textwidth]{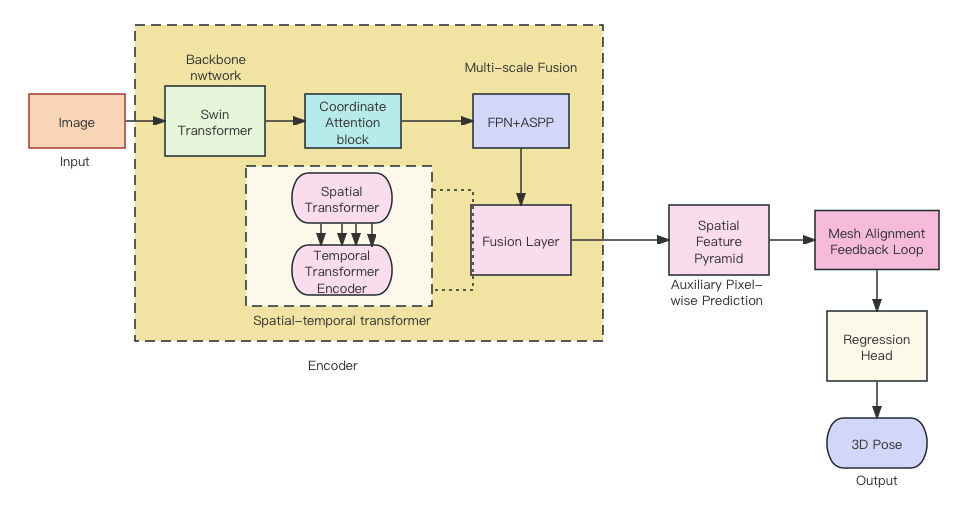}
    \caption{The complete architecture of the proposed PyCAT4 model, integrating a Swin Transformer backbone, Coordinate Attention, Multi-scale Fusion (FPN+ASPP), and a Spatial-Temporal Fusion module, followed by the original PyMAF prediction heads.}
    \label{fig:pycat4_arch}
\end{figure*}

The final PyCAT4 architecture, shown in Fig.~\ref{fig:pycat4_arch}, is a unified framework that synergistically combines the aforementioned enhancements.

\subsubsection{Architectural Innovations}
PyCAT4 integrates the following key technologies:
\begin{itemize}
    \item \textbf{Coordinate Attention (CA) Mechanism:} Enhances focus on relevant spatial features crucial for accurate joint detection.
    \item \textbf{Swin-Transformer Backbone:} Replaces the traditional CNN to efficiently model long-range dependencies and hierarchical features.
    \item \textbf{Multi-Scale Feature Fusion:} Utilizes a combination of FPN and ASPP for a comprehensive analysis of both global and local features.
    \item \textbf{Temporal Fusion Module:} Employs a Transformer-based approach to synthesize information from consecutive video frames, improving dynamic pose estimation.
\end{itemize}

\subsubsection{Unified Architecture Description}
The data flow in PyCAT4 is designed for cohesive operation:
\begin{enumerate}
    \item An input image or video sequence is fed into the Swin-Transformer backbone to extract a rich set of hierarchical features.
    \item These features are refined by the Coordinate Attention module to emphasize important spatial locations.
    \item The FPN and ASPP structures combine features from various scales to ensure no detail is overlooked.
    \item For video inputs, the temporal fusion module integrates these features across time to stabilize pose estimation.
    \item Finally, the processed features are mapped to precise 3D pose and shape predictions by the regression head.
\end{enumerate}

\subsection{Testing Environment}
All experiments were conducted on a high-performance server with the configuration detailed in Table~\ref{tab:config}.

\begin{table}[!t]
    \caption{Hardware and Software Configuration for Experiments}
    \label{tab:config}
    \centering
    \begin{tabular}{|l|l|}
        \hline
        \textbf{Component} & \textbf{Specification} \\
        \hline
        Operating System & Ubuntu 20.04 \\
        \hline
        GPUs & 6 x NVIDIA RTX 4090 (24GB) \\
        \hline
        CPU & 32 vCPU Intel(R) Xeon(R) Gold 6430 \\
        \hline
        Memory & 240GB RAM \\
        \hline
        Software Stack & PyTorch 1.10.0, Python 3.8, CUDA 11.3 \\
        \hline
    \end{tabular}
\end{table}

\subsection{Loss Configuration}
The training objective includes multiple loss components: a classification loss for foreground-background segmentation, a regression loss for keypoints, and an error loss for camera parameters. We adopt the IUV map representation from DensePose, applying cross-entropy loss for partial index classification and regression loss for UV coordinates across three feature scales.

\subsection{Benchmarking and Model Modifications}
The project utilizes the original PyMAF model as a baseline, with the aforementioned modifications introduced incrementally. Benchmark tests are performed on the COCO and 3DPW datasets to systematically compare the performance of each model variation.


\section{Experiments and Results}
This section presents the comprehensive evaluation of our proposed PyCAT4 model. We detail the data preprocessing pipeline, present quantitative results from extensive ablation studies on the 3DPW and COCO datasets, provide qualitative visual comparisons, and discuss the implications of our findings.

\subsection{Data Preprocessing and Augmentation}
Our experiments utilize a composite dataset created by merging the 3DPW and COCO datasets, comprising approximately 70,000 images. The 3DPW dataset provides annotations for 3D human poses, while the COCO dataset offers annotations for 2D human keypoints. To enable mixed training, we harmonized both datasets into a unified format. To enhance model robustness and mimic the variability of real-world scenarios, we applied standard data augmentation techniques, including random cropping, scaling, and rotation, during the training process.

\subsection{Ablation Studies}
To systematically validate the contribution of each component in our proposed architecture, we conducted a series of ablation studies. Starting with the original PyMAF model as our \textbf{Baseline}, we incrementally integrated the following enhancements:
\begin{itemize}
    \item \textbf{CA Model:} The baseline augmented with a \textbf{C}oordinate \textbf{A}ttention mechanism.
    \item \textbf{CA\_Transformer Model:} Integrates a \textbf{Swin-Transformer} backbone after the CA module.
    \item \textbf{CA\_FPN\_Transformer Model:} Adds a multi-scale fusion framework (\textbf{F}eature \textbf{P}yramid \textbf{N}etwork) to the previous model.
    \item \textbf{PyCAT4 (CA\_FPN\_Transformer\_4D):} The final model, which incorporates a temporal fusion component for \textbf{4D} (3D + time) analysis.
\end{itemize}
The performance of each model was assessed on the validation sets of 3DPW and COCO.

\subsubsection{3D Pose Estimation Results on 3DPW}
For 3D human mesh recovery, we use three standard metrics: Mean Per-Vertex Error (PVE), Mean Per-Joint Position Error (MPJPE), and Procrustes-Aligned MPJPE (PA-MPJPE). All metrics are reported in millimeters (mm), where lower values indicate better performance. Table~\ref{tab:3dpw_results} summarizes the performance on the 3DPW validation set.

\begin{table}[!t]
    \caption{Performance Comparison on the 3DPW Validation Set. The Percentage Improvement is Relative to the Baseline. Lower is Better ($\downarrow$).}
    \label{tab:3dpw_results}
    \centering
    \resizebox{\columnwidth}{!}{%
    \begin{tabular}{l|ccc|c}
        \hline
        \textbf{Model} & \textbf{PVE} ($\downarrow$) & \textbf{MPJPE} ($\downarrow$) & \textbf{Recon. Error} ($\downarrow$) & \textbf{PVE Improv. (\%)} \\
        \hline
        Baseline & 134.11 & 115.89 & 67.42 & - \\
        CA & 133.45 & 115.40 & 67.43 & -0.49\% \\
        CA\_Transformer & 111.00 & 97.08 & 57.68 & -17.23\% \\
        CA\_FPN\_Transformer & 111.38 & 97.93 & 57.60 & -16.95\% \\
        \textbf{PyCAT4 (Ours)} & \textbf{108.50} & \textbf{92.73} & \textbf{55.98} & \textbf{-19.10\%} \\
        \hline
    \end{tabular}}
\end{table}

\subsubsection{2D Pose Estimation Results on COCO}
For 2D human pose estimation, we evaluate using Average Precision (AP) and Average Recall (AR) based on Object Keypoint Similarity (OKS). Higher values indicate better performance. Table~\ref{tab:coco_results} shows the results on the COCO validation set.

\begin{table*}[!t]
    \caption{Performance Comparison on the COCO Validation Set. All Metrics are Higher is Better ($\uparrow$).}
    \label{tab:coco_results}
    \centering
    \begin{tabular}{l|ccc|ccc}
        \hline
        \textbf{Model} & \textbf{AP@50:95} ($\uparrow$) & \textbf{AP@50} ($\uparrow$) & \textbf{AP@75} ($\uparrow$) & \textbf{AR@50:95} ($\uparrow$) & \textbf{AR@50} ($\uparrow$) & \textbf{AR@75} ($\uparrow$) \\
        \hline
        Baseline & 0.135 & 0.392 & 0.058 & 0.273 & 0.599 & 0.213 \\
        CA & 0.152 & 0.424 & 0.075 & 0.294 & 0.620 & 0.243 \\
        CA\_Transformer & 0.292 & 0.610 & 0.239 & 0.439 & 0.753 & 0.454 \\
        CA\_FPN\_Transformer & 0.292 & 0.613 & 0.245 & 0.448 & 0.766 & 0.464 \\
        \textbf{PyCAT4 (Ours)} & \textbf{0.295} & \textbf{0.621} & \textbf{0.246} & \textbf{0.449} & \textbf{0.766} & \textbf{0.466} \\
        \hline
        \textbf{Improvement (\%)} & \textbf{+118.52\%} & \textbf{+58.42\%} & \textbf{+324.14\%} & \textbf{+64.47\%} & \textbf{+27.88\%} & \textbf{+118.78\%} \\
        \hline
    \end{tabular}
\end{table*}

\subsection{Qualitative Comparison}
In addition to quantitative metrics, we provide visual comparisons to intuitively demonstrate the superiority of PyCAT4.

\subsubsection{Single Image Reconstruction}
\begin{figure}[!t]
    \centering
    \includegraphics[width=\columnwidth]{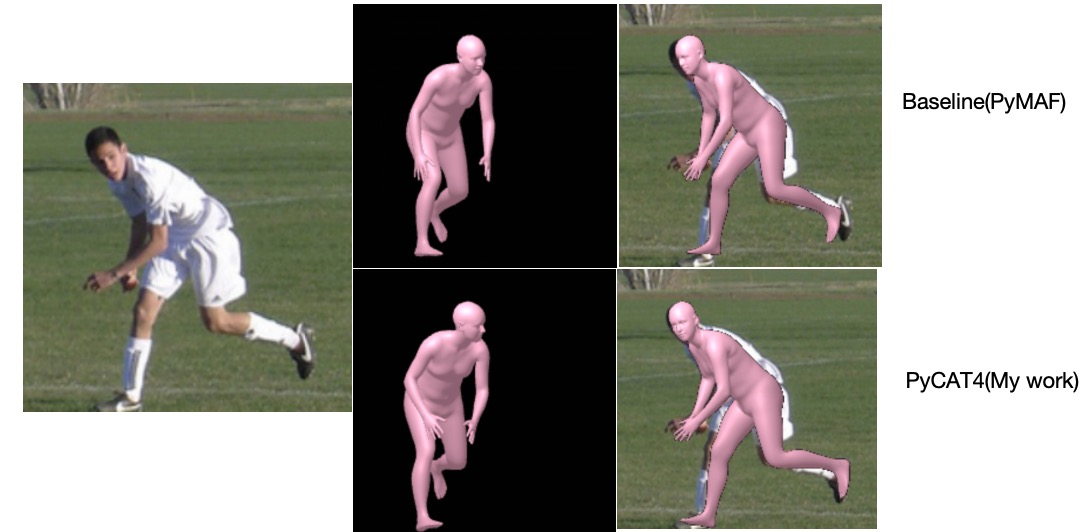}
    \caption{Visual comparison for single-image 3D reconstruction. PyCAT4 (bottom row) demonstrates a more accurate mesh alignment, especially on the torso and limbs, compared to the baseline PyMAF (top row).}
    \label{fig:visual_image}
\end{figure}

As shown in Fig.~\ref{fig:visual_image}, PyCAT4 achieves a significantly better fit to the person in the image compared to the baseline model. The improvements are particularly noticeable in complex areas like the hands, feet, and face, where our model captures joint dynamics and body shape with greater accuracy.

\subsubsection{Multi-person Video Reconstruction}
\begin{figure}[!t]
    \centering
    \includegraphics[width=\columnwidth]{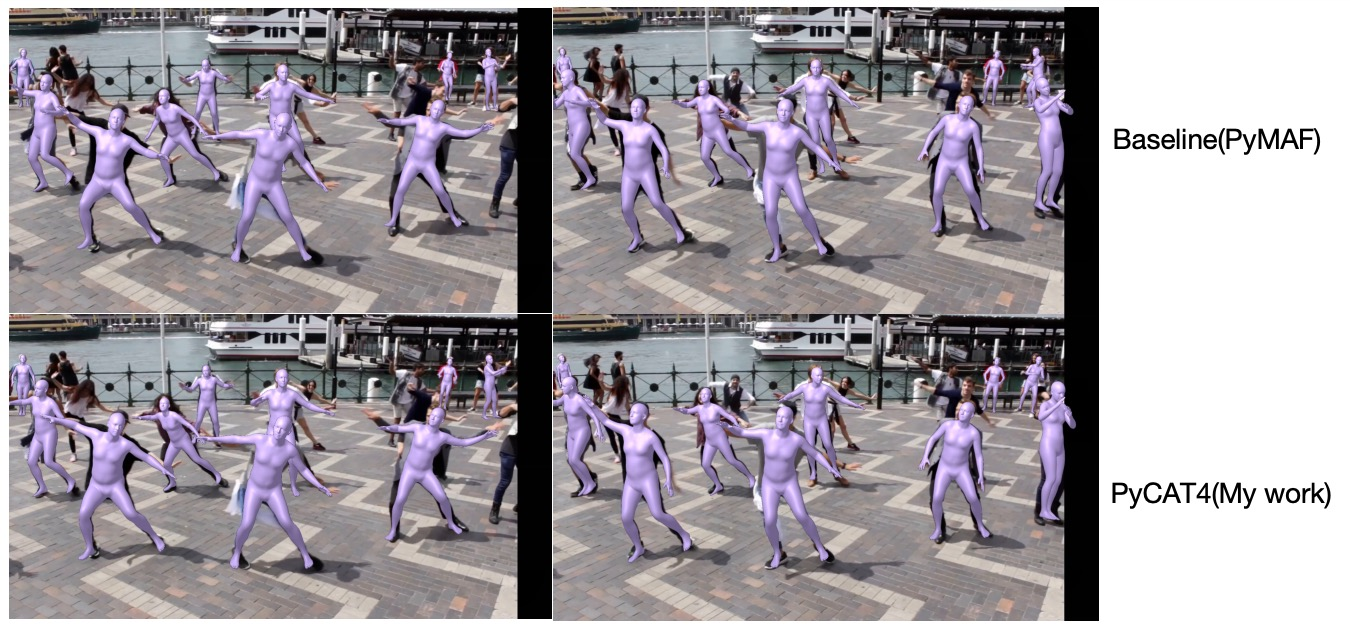}
    \caption{Qualitative comparison for multi-person video. PyCAT4 (bottom) provides more reliable and precise pose estimations in crowded scenes than the baseline (top).}
    \label{fig:visual_video}
\end{figure}

For video inputs, PyCAT4 not only improves visual quality (Fig.~\ref{fig:visual_video}) but also enhances processing speed. As detailed in Table~\ref{tab:timing}, our model reduces the time required for both multi-person tracking and per-tracklet reconstruction, making it more suitable for real-time or near-real-time applications. The improved fitting accuracy, especially in detailed regions, is attributable to the integration of the CA and Swin-Transformer modules, which allow for more effective processing of spatial context.

\begin{table}[!t]
    \caption{Processing Time Comparison for a 200-Frame Video Clip.}
    \label{tab:timing}
    \centering
    \begin{tabular}{l|cc}
        \hline
        \textbf{Task} & \textbf{Baseline (PyMAF)} & \textbf{PyCAT4 (Ours)} \\
        \hline
        Multi-person Tracker & 17s & \textbf{14s} \\
        Reconstruction & 3m 27s & \textbf{3m 1s} \\
        Rendering Output & 12m 58s & 12m 58s \\
        \hline
    \end{tabular}
\end{table}

\subsection{Discussion}
Our experimental results provide several key insights. The integration of the \textbf{Coordinate Attention (CA)} mechanism yielded significant gains on the 2D COCO dataset (e.g., a 29.31\% improvement in AP@75), indicating its effectiveness in focusing on key regions for 2D feature extraction.

Replacing the backbone with the \textbf{Swin-Transformer} brought substantial improvements across both 2D and 3D datasets, highlighting its superior capability in modeling global dependencies and extracting more balanced deep features compared to a traditional CNN.

The introduction of \textbf{multi-scale fusion} and, particularly, \textbf{temporal fusion} led to the largest performance gains on the 3D 3DPW dataset. The temporal module's ability to leverage information across consecutive frames is crucial for accurate dynamic pose estimation.

Overall, the final \textbf{PyCAT4} model, which integrates all proposed enhancements, achieved the best performance across most metrics. This demonstrates that a hybrid strategy, combining a powerful Transformer backbone with specialized attention, multi-scale, and temporal modules, provides a robust and effective solution for complex human pose estimation tasks.

\subsection{Model Validation Techniques}
To ensure a fair and robust assessment of the different network architectures, we employed a strict control variable approach throughout our ablation studies. Uniform training parameters—including the number of epochs, learning rate, and batch size—were maintained across all models. This controlled environment ensures that observed performance differences can be directly attributed to the specific architectural modifications rather than variations in training conditions. Each model configuration was tested multiple times, and the results were averaged to mitigate the effects of anomalous training runs, thereby ensuring the reliability of our findings.

\subsection{Real-time Testing Methodology}
\begin{figure}[!t]
    \centering
    \subfloat[Live output window]{\includegraphics[width=1.5in]{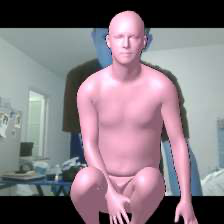}\label{fig:realtime_output}}
    \hfil
    \subfloat[System in operation]{\includegraphics[width=1.7in]{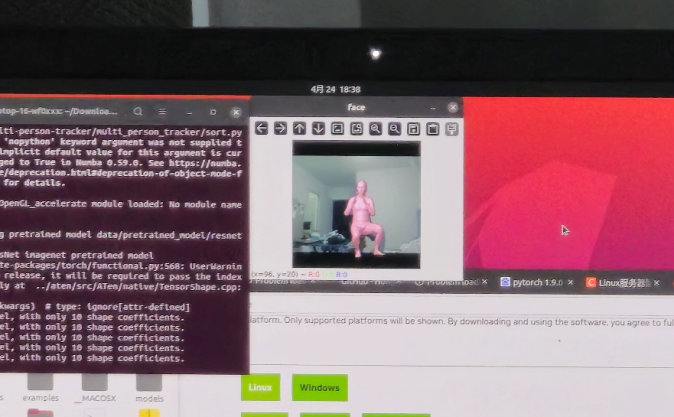}\label{fig:realtime_system}}
    \caption{Demonstration of the real-time 3D pose estimation system using the PyCAT4 model and OpenCV.}
    \label{fig:realtime_demo}
\end{figure}

We utilized the OpenCV library to develop and test a real-time inference system. The trained PyCAT4 model was integrated into an OpenCV framework that captures live video from a camera, processes each frame on a GPU for real-time inference, and subsequently displays the 3D mesh overlay in a window. This demonstrates the practical capability of our system for real-time applications, as shown in Fig.~\ref{fig:realtime_demo}.

\subsection{Limitations}
Our study has two primary limitations. First, the \textbf{comparison scope} was confined to our baseline, PyMAF. The absence of comparisons against a broader range of state-of-the-art architectures limits the generalizability of our performance claims. Second, \textbf{dataset restrictions} prevented us from using the Human3.6M dataset due to licensing issues. Consequently, our study was confined to the 3DPW and COCO datasets, which may not fully reflect the performance that could be achieved with a more comprehensive training regimen.


\section{Conclusion and Future Work}
In this paper, we presented PyCAT4, a novel framework for 3D human pose and shape estimation that significantly advances the state-of-the-art over baseline models. Our approach successfully integrates a \textbf{Swin-Transformer} backbone for powerful hierarchical feature extraction, a \textbf{Coordinate Attention (CA)} module to focus on critical image regions, \textbf{spatial pyramid structures} for robust multi-scale feature fusion, and a \textbf{temporal fusion} component to effectively analyze motion dynamics in video sequences. Rigorous experiments on the COCO and 3DPW datasets validated the effectiveness of these integrated technologies, demonstrating substantial improvements across all key evaluation metrics. The development of a functional real-time system further underscores the practical applicability of our work.

While PyCAT4 demonstrates significant progress, the field of human pose estimation continues to present challenges, particularly in handling extreme motion complexity and improving computational efficiency for real-time applications. This work also highlights the critical importance of ethical considerations in deploying such technologies.

\subsection{Ethical Considerations}
The deployment of advanced pose estimation technologies necessitates careful consideration of several ethical issues to ensure responsible usage:
\begin{itemize}
    \item \textbf{Privacy Concerns:} Pose estimation can process sensitive personal data. It is crucial to implement data anonymization and comply with privacy regulations, obtaining informed consent wherever possible.
    \item \textbf{Data Security:} Robust measures, including encryption and strict access controls, are vital to protect data during transmission and storage, preventing potential breaches.
    \item \textbf{Bias and Fairness:} Models may amplify biases present in training data. Employing diverse datasets and conducting regular audits are essential to ensure fairness and prevent discriminatory outcomes.
    \item \textbf{Transparency and Accountability:} The deployment and purpose of pose estimation technologies must be transparent. Organizations should be accountable for their ethical use.
    \item \textbf{Accessibility:} Technologies should be designed to be inclusive and accommodate a wide range of human movements and abilities, avoiding discrimination against individuals with disabilities.
\end{itemize}

\subsection{Future Work}
Based on the outcomes of this project, we identify several promising directions for future research:
\begin{itemize}
    \item \textbf{Multitask Learning Frameworks:} Integrating pose estimation with related tasks, such as action recognition or emotion detection, could lead to more comprehensive models that leverage shared features for improved overall performance.
    \item \textbf{Semi-Supervised Learning:} Exploring semi-supervised methods can help address the scarcity of labeled 3D data by effectively leveraging large amounts of unlabeled video, potentially improving model generalization.
    \item \textbf{Dataset Diversity:} Incorporating a wider range of datasets featuring more diverse demographics, environments, and activities will be crucial for enhancing the robustness and real-world applicability of pose estimation models.
    \item \textbf{Real-Time System Enhancements:} Further optimizing the model architecture for efficiency, possibly through model compression or quantization, and leveraging hardware-specific accelerations can improve real-time performance on edge devices.
    \item \textbf{Cross-Modal Learning:} Exploring the fusion of data from additional modalities, such as depth sensors or audio cues, could significantly enhance predictive accuracy, especially in complex or ambiguous scenarios where visual data alone is insufficient.
\end{itemize}
By addressing these areas, future work can continue to push the boundaries of human pose estimation, making systems more accurate, efficient, and applicable to a wider range of real-world challenges.

\section*{Acknowledgment}

The author wishes to thank his supervisor, Jonathan Loo, for his invaluable guidance and support throughout this research project. The author also acknowledges the developers of the open-source libraries and datasets that made this work possible.


\bibliographystyle{IEEEtran}
\bibliography{references}


\section{Biography Section}
\begin{IEEEbiographynophoto}{Zongyou Yang}
is currently pursuing the M.Sc. degree in Computer Graphics, Vision and Imaging at University College London (UCL). He received the B.Sc. (Eng.) degree with First Class Honours in telecommunications engineering with management from Queen Mary University of London, in a joint program with the Beijing University of Posts and Telecommunications (BUPT).

His research interests include Machine Learning, Deep Learning, Computer Vision, and Data Science.

\end{IEEEbiographynophoto}

\vfill

\end{document}